\documentclass{article} 
\usepackage{nips14submit_e,times}
\usepackage{hyperref}
\usepackage{url}
\usepackage{amsmath,amssymb}
\usepackage[noend,linesnumbered,noline,ruled,titlenotnumbered]{algorithm2e}
\usepackage{enumitem}
\usepackage{bbm}
\usepackage{verbatim} 
\usepackage{graphicx}
\usepackage{epstopdf}
\usepackage{subfig}
\usepackage{mathrsfs}

\newtheorem{theorem}{Theorem}
\newtheorem{cor}{Corollary}
\newtheorem{defn}{Definition}

\title{Lifted Message Passing for the\\ Generalized Belief Propagation}

\author{
Udi Apsel \\
Department of Computer Science\\
Ben Gurion University of The Negev, Israel\\
\texttt{apsel@cs.bgu.ac.il}
}

\newcommand{\GR}{G_{\mathcal{R}}}

\newcommand{\xI}{x_{\alpha}}
\newcommand{\xIt}{x_{\alpha'}}
\newcommand{\xJ}{x_{\mathcal{\beta}}}

\nipsfinalcopy 

\begin{document}

\maketitle

\begin{abstract}
We introduce the lifted Generalized Belief Propagation (GBP) message passing algorithm, for the computation of sum-product queries in Probabilistic Relational Models (e.g.~Markov logic network). The algorithm forms a compact region graph and establishes a modified version of message passing, which mimics the GBP behavior in a corresponding ground model. The compact graph is obtained by exploiting a graphical representation of clusters, which reduces cluster symmetry detection to isomorphism tests on small local graphs. The framework is thus capable of handling complex models, while remaining domain-size independent.

\end{abstract}

\section{Introduction}
Probabilistic Relational Models (PRM) (e.g.~Markov logic network \cite{DBLP:journals/ml/RichardsonD06}) are compact and expressive representations of probabilistic models, which succinctly capture probabilistic rules using the language of \textit{first-order predicate logic}. Albeit their compactness, inferring from these rules is a challenging task, which gave rise to a family of algorithms bundled under the name \textit{lifted inference}, dedicated to exploiting the inherent symmetry exhibited by the compact representations. One of the popular lifted inference methods is an adaptation of the famous sum-product Belief-Propagation (BP) \cite{yedidia2003understanding} algorithm to relational models \cite{kersting09uai,singla08aaai}. Based on a synchronous message passing schedule which exploits the symmetry of the relational model, lifted BP manages to compress huge probabilistic models into surprisingly small representations, while mimicking the BP behavior exactly. 

In this paper we introduce the first domain-size independent framework which lifts the generalized BP algorithm (GBP) \cite{yedidia2003understanding} in its classical message passing form, and thus allows the injection of more constraints on the marginals compared with non-generalized BP implementations. A related work by \cite{DBLP:conf/uai/BroeckCD12} introduced a method which produces similar approximations, by relaxing the relational model's structure \cite{DBLP:conf/aaai/ChoiD06}, compensating for the relaxation and finally performing exact inference. Their implementation shows good results on many instances, however the method's execution time is still polynomial in the domain size. Our method, in contrast, is entirely domain-size independent (in case of no evidence), and does not rely on any external engine for inference.

Our work heavily relies on a recently introduced graphical platform called Cluster Signature Graph (CSG) \cite{apsel14AAAI}, which projects the relational structure of clusters of variables onto a graph, and allows symmetry detection via an isomorphism test. Based on this platform, we formulate a compact representation of the \textit{region graph}, which is the graphical structure used for the GBP message passing. This \textit{lifted} region graph is accompanied by a modified version of message passing, which mimics the GBP behavior in a respective ground model. The core reliance on a graphical representation enables us to frame most parts of this work in graphical terms, which are sometimes separate from terms used in similar lifted inference works. Nevertheless, this high-level perspective is what makes the framework capable of handling relational models of complex structure.

We begin with two background section, one introducing the GBP algorithm and related concepts, the other providing background on relational models and the CSG platform. The lifted GBP framework is presented next, starting with an overview, and continuing with the more formal parts of this work. We conclude with an empirical demonstration of the framework and a brief discussion.

\section {Background}
\subsection {Inference in Markov Random Field (MRF)}
A Markov Random Field (\textit{MRF}) $\mathcal{M}$  is a probabilistic graphical model, consisting of a set of random variables $x=\{x_i\}_i$ and a set of \textit{factors}. A factor is a pair $(\phi_f,x_f)$, which represents a function $\phi_f:\mathrm{range}(x_f) \rightarrow \mathbb{R}^+$, mapping from the joint assignment range of variables $x_f \subseteq x$, to the non-negative reals. The joint distribution function of MRFs is given by
\begin{equation}
\label{label:joint_prob}
Pr(x = \hat{x}) = \frac{1}{Z} \prod\nolimits_f { \phi_f( \hat{x}_f )  } 
\end{equation}
where $\hat{x}$ is a joint assignment to all $x$ variables, $\hat{x}_f$ is the respective joint assignment to all $x_f$ variables under $\hat{x}$, and $Z$ denotes a normalization constant called the \textit{partition function}. A common task in MRFs is that of \textit{marginalization}, which is computing the probability of all possible states in a subset of variables $x' \subseteq x$ , as follows.
\begin{equation}
\label{label:MARG}
Pr(x')  \propto \sum_{x \setminus x'} \prod\nolimits_f {\phi_f( x_f) } 
\end{equation}
The result of Equation \ref{label:MARG} is a function, mapping from $\mathrm{range}(x')$ to the non-negative reals.

\subsection {Generalized Belief Propagation (GBP)}
The marginalization task is \#P-complete \cite{DBLP:journals/ai/Roth96}, and it is therefore common to approximate its result, rather than to carry out exact computations. One such approximation method is the sum-product Belief Propagation (BP). The algorithm schedules messages between neighboring nodes in the graphical models, until messages converge, and thus simultaneously computes marginals on all random variables in the model. Although BP does not guarantee to converge in graphs that contain loops, the procedure often arrives at a reasonable set of approximations to the correct marginal distributions \cite{DBLP:journals/jmlr/IhlerFW05}. The result then corresponds to a stationary point of the Bethe free energy approximation. 

The Generalized Belief Propagation (GBP) algorithm is, as its name suggests, a generalization of the BP algorithm. GBP messages are sent from one cluster of variables to another, in a graphical structure called a region graph. When the algorithm converges, the result corresponds to a stationary point of the Kikuchi free energy approximation \cite{yedidia2003understanding}, a tighter free energy approximation compared with Bethe. In this this work we focus on the parent to child message passing variation of GBP.

\subsubsection {Regions and Region Graphs}
A \textit{region} \cite{DBLP:conf/uai/Welling04} is a tuple $(\xI,f_\alpha)$, which represents a node in the GBP message passing graph. $\xI$ is a cluster of MRF variables with a respective set of indices $\alpha$, and $f_\alpha$ is a set of MRF factors whose scope is a subset of $\xI$ (or $\xI$ entirely). For simplicity and notational convenience, we will assume that $f_\alpha$ contains all factors under the $\subseteq \xI$ scope. Hence, the notation $\alpha$ will be sufficient to denote a region of a corresponding scope of variables.

A \textit{region graph} $G_{\mathcal R} = (V,E)$ is a Directed Acyclic Graph, with nodes in $V$ denoting regions, and directed edges in $E$ denoting parent to child (source to target) relations. We define the conditions which a region graph must respect, as follows. (1) For every pair of distinct regions $\alpha_1,\alpha_2$ which are not subsets of one another, there exists an \textit{intersection region} $\beta = \alpha_1 \cap \alpha_2$. (2) For every pair of regions $\alpha,\beta$,  the proposition $\beta \subset \alpha$ is true iff $\beta$ is a descendant of $\alpha$; (3) The set of parent-less regions (called \textit{outer regions}) must consist of all scopes of MRF factors.

Generating region graphs can be understood as an iterative process \cite{yedidia2003understanding}, where region intersections are applied, first on outer regions, which are given as input, and then on the resulting intersections. For example, given the set of outer regions $\{1,2,3,4\}, \{1,2,5,6\}, \{1,3,5,7\}$, a region graph is generated such that intermediate intersections $\{1,2\},\{1,3\}$ and $\{1,5\}$ are added, and a subsequent intersection $\{1\}$ is added as a child to all these intermediate intersections. 

\subsubsection {Parent-To-Child Message Passing}
Let $(\xI, f_\alpha)$ denote a region in the region graph $\GR$. We define $Par(\alpha)$ as the set of all regions that are parents of $\alpha$, and $Desc(\alpha)$ as the set of all its descendants. The algorithm starts by arbitrarily defining messages $m_{\alpha \rightarrow \beta}$ from all parent regions to their child regions. At each phase of the algorithm, messages are updated according to the following rules.
\begin{equation}
b_\alpha (\xI) \propto \prod_{f \in f_\alpha}{\phi_f(x_f)} \prod_{\rho \in Par(\alpha)} {m_{\rho \rightarrow \alpha}(\xI)}
\prod_{\gamma \in Desc(\alpha)} \; \prod_{\rho \in Par(\gamma) \setminus \{\alpha\} \setminus Desc(\alpha)}{m_{\rho \rightarrow \gamma}(x_\gamma)}
\end{equation}
\begin{equation}
m_{\alpha \rightarrow \beta}(x_\beta) \longleftarrow \frac{\sum_{x_\alpha \setminus x_\beta} {b_\alpha(x_\alpha)} } {b_\beta(x_\beta)} \; m_{\alpha \rightarrow \beta}(x_\beta)
\end{equation}
When the algorithm converges, the belief state of $\xI$ is obtained via the computation of $b_{\alpha}(\xI)$.

\section {Probabilistic Relational Models}
Probabilistic Relational Models (PRM) are representations of probabilistic models using the language of first-order predicate logic. Of the two  most common models, \textit{Markov logic network} \cite{DBLP:journals/ml/RichardsonD06} and the \textit{parfactor model} \cite{ DBLP:conf/ijcai/Poole03}, we choose the latter to represent a relational MRF. We thus include a brief introduction to the parfactor model, and refer to \cite{DBLP:conf/ijcai/BrazAR05} for a more comprehensive overview.

\subsection{Relational MRF} 
A \textit{domain} is a set of constants, called \textit{domain objects}, that represent distinctive entities in the modeled world, e.g.~$\{Alice, Bob, Carol, \ldots\}$. A logical variable (\textit{lvar}) is a variable whose assignment range is associated with some domain. An \textit{atom} is an atomic formula of the form $p(t_1,\ldots,t_n)$, where the symbol $p$ is called a \textit{predicate}\footnotemark, and each term $t_i$ is either a domain object or an lvar. A \textit{ground atom} is an atom whose $t_i$ terms are all domain objects. Non-ground atoms are collections of ground atoms, all sharing the same assignment range, and describing a certain property of an individual (e.g.~smoker) or some relation between individuals (e.g.~friendship). A \textit{ground substitution} $\{X_i/o_i\}_i$, is the replacement of each lvar $X_i$ with a domain object $o_i$.
\footnotetext[1]{Although the term predicate is used, atoms are not restricted to Boolean assignments.}

The parfactor model $\mathcal{M}^r$ (aka \textit {relational MRF}) is a collection of relational factors, called \textit{parfactors}. A parfactor is a tuple $(\phi, A, R)$, consisting of a function $\phi:\mathrm{range}(A) \rightarrow \mathbb{R}^+$, an ordered set of atoms $A$, and a set of constraints $R$ imposed on $A$'s lvars. Grounding a parfactor is done by applying all ground substitutions that are consistent with $R$, resulting in a collection of factors.  The ground atoms then serve as random variables in the ground MRF. A notation $\phi(A \hspace{2pt} | \hspace{2pt}R)$ is commonly used to denote a parfactor. For example, parfactor $\phi( sm(X), sm(Y), fr(X,Y)  | X \neq Y)$, whose ground instances in the domain $\{Alice,Bob\}$ are (i) $\phi \big( sm(Alice), sm(Bob), fr(Alice,Bob) \big)$ and (ii) $\phi \big( sm(Bob), sm(Alice), fr(Bob,Alice) \big)$.

We restrict our attention to shattered \cite{DBLP:conf/ijcai/BrazAR05} models, consisting of inequality constraints of the form $X \neq Y$ only. Additionally, such inequality constraints will be imposed on each pair of lvars $X,Y$ where, in their absence, a ground factor with multiple entries of the same ground atom may be produced. For instance, $\phi(p(X),p(Y))$ may produce a ground factor $\phi(p(o_1),p(o_1))$, and will therefore be split into two parfactors:  $\phi(p(X),p(Y) | X \neq Y)$ and some $\phi'(p(X))$. Finally, the notations $r_{(i,j)}$ and $p_i$ will be used to abbreviate $r(o_i,o_j)$ and $p(o_i)$, respectively.

\subsection {Symmetry Between Clusters}
The first-order representation of relational models introduces a substantial amount of symmetry, which can be exploited for either exact or approximate inference. In exact inference, computational operators must typically take into account the partitioning of the model into isomorphic components \cite{DBLP:conf/nips/TaghipourDB13}. In comparison, approximate inference methods \cite{niepert12uai,Bui13,mladenov14aistats} are able to exploit a more relaxed form of symmetry, one that is exhibited between clusters of MRF variables. 
\begin{defn}
Clusters $\xI$ and $\xIt$ are said to be symmetrical if there exists a structure preserving permutation $\pi$ on the MRF's variables (i.e.~a permutation belonging to the automorphism group of the graphical model), under which each $x_i \in \xI$ is mapped onto a distinctive $x_j \in \xIt$. 
\end{defn}

In \cite{apsel14AAAI}, a platform based on graphical signatures of clusters is introduced, providing a principled way to incorporate clustering  based methods in lifted inference.
\subsubsection{Cluster Signatures and Canonical Clusters}
Let $\xI$ denote a cluster of ground atoms obtained from the relational MRF $\mathcal{M}^r$.  The Cluster Signature Graph (CSG) of $\xI$ is the projection of its content onto a graph, in a way that guarantees two important properties : (i) If the CSGs of $\xI$ and $\xIt$ are isomorphic, then $\xI$ and $\xIt$ are symmetrical. (ii) The mapping induced by such an isomorphism constitutes a structure preserving permutation in the relational MRF. For lack of space, we present a simpler definition of CSG than the one introduced by \cite{apsel14AAAI}, which captures slightly less symmetry and pertains to shattered models.
\begin {defn}
The CSG of cluster $\xI$ is a directed colored multigraph $G = (V, E,C)$, where $V$ is a set of vertices, $E$ is a set of directed edges and $C$ is a coloring function, mapping each edge to a color. Edges and colors in the CSG are defined as follows. (1) For each member of $\xI$ originating from a unary ground atom $p_i$, let $G$ contain a node $i$ and a self-edge carrying the color '$p$'; (2) For each member of $\xI$ originating from a binary ground atom $r_{(i,j)}$, let $G$ contain the nodes $i$ and $j$ and a directed edge $i \rightarrow j$ carrying the color '$r$'.
\end{defn} 
One important feature of CSGs is that of \textit{canonical clusters}. Canonical clusters are unique representatives for all clusters belonging to the same symmetry class. Each cluster $\xI$ corresponds to a canonical representative $x_{\alpha^*}$, that can be obtained by applying \textit{canonical labeling} to the CSG of $\xI$ using dedicated graph canonization tools (e.g.~\textit{nauty} \cite{McKay201494}), and extracting the canonical cluster's variable members from the edges of the canonically labeled graph.

Since CSGs consist only of cluster members, all related tasks (canonization, isomorphism tests, etc.) which are derived from this graphical representation are, at worst, exponential in the cluster size. This locality property is what makes the framework efficient and domain-size independent. Although CSGs are also capable of detecting symmetry in presence of evidence, we avoid this more complicated representation in this work, and will simply resort to the shattering of the model.

\section {Lifting the Generalized Belief Propagation}
Lifting the generalized belief propagation can be regarded as a two stage process. First, given a relational MRF and a set of outer regions, a compact region graph is formed, entirely comprised of regions corresponding to canonical clusters. Second, a message passing algorithm, specifically adapted to the lifted graph, is applied, producing the exact same result as would the ground GBP. We begin with a short overview of the lifted framework.

\subsection {Lifted GBP -- An Overview}
\label{subsec:overview}
Consider obtaining a region graph for the relational MRF $\phi(p(X),q(Y) | X \neq Y)$, where all pairs $\{p_i,q_j\}$ ($i \neq j$) serve as outer regions. Such a graph can be obtained by finding all the intersections induced by the $\{p_i,q_j\}$ pairs, and defining two edges from each pair to its members:  $\{p_i,q_j\} \rightarrow p_i$ and $\{p_i,q_j\} \rightarrow q_j$. However, the highly symmetric nature of the region graph allows to capture its structure much more succinctly, via a representation which simulates the description given above.

We first define symmetry in the context of region graphs. Regions $\alpha$ and $\alpha'$ are said to be symmetrical if there exists a permutation $\pi_{_{\mathcal{R}}}$ on the region graph's nodes which preserves its structure, and under which $\pi_{_{\mathcal{R}}}(\alpha) = \alpha'$. Somewhat unsurprisingly, symmetry of regions is directly derived from the symmetry of their respective clusters. Here, all pairs $\{p_i,q_j\}$ are symmetrical in the relational model, and thus so are the corresponding regions. The same applies to all the atomic $p$ and $q$ regions.

At this point, we would like to utilize the canonical representation of clusters to form a compact (lifted) region graph, comprised entirely of such clusters. Consider the cluster $\{p_i,q_j\}$, whose region will serve as a canonical representative for all pairs. Similarly, let $p_k$ and $q_m$ denote the canonical representatives for the single atom regions. The lifted region graph should then consist of two edges: $\{p_i,q_j\} \rightarrow p_k$ and $\{p_i,q_j\} \rightarrow q_m$. To represent the flow of messages from parent to child, each edge is accompanied by a mapping, expressing which "role" the child assumes w.r.t.~the parent members. Here, the mapping $p_i \mapsto p_k$ will be associated with one edge, and $q_j \mapsto q_m$ will be associated with the other. Still, edges in the lifted graph must indicate how many symmetrical parents are connected to a single child, in the role specified by each edge. In our example, this  \textit{cardinality} is quite natural. If the domain size is $N$, then each $p_k$ is connected to $N-1$ parents from the $\{p_i,q_j\}$ symmetry class under then role of $p_i$, and each $q_m$ is connected to $N-1$ such parents under then role of $q_j$.

Consider now a slightly different model, consisting entirely of $\{p_i,p_j\}$ pairs. Let all pairs $\{p_i,p_j\}$ and all single $p_k$ instances be symmetrical. The lifted region graph must then consist of two nodes,  $\{p_i,p_j\}$ and $p_k$, to denote the canonical representatives as before, but instead of two edges from $\{p_i,p_j\}$ to $p_k$, only one edge is required. The reason, which will be formalized in the next subsection, is that the special structure of the model forces messages being sent from each $\{p_i,p_j\}$ to $p_i$, to be entirely identical to those sent to $p_j$. $p_k$ can then assume one of the roles $p_i$ or $p_j$, arbitrarily, and still reflect correctly the message flow in the ground region graph.

Lastly, we wish to simulate message passing in the ground graph by sending messages in the lifted graph. Such simulation is possible if the messages in the ground graph are scheduled such that the graphical model's symmetry is reflected in the messages. This is indeed the case with the \textit{flooding} schedule where, at each iteration, all nodes send messages at the exact same time. Under an assumption of initial message symmetry, if regions $\alpha$ and $\beta$ are mapped to $\alpha'$ and $\beta'$ by a structure preserving permutation, then at any iteration of the algorithm, messages from $\alpha$ to $\beta$ remain symmetrical to those sent from $\alpha'$ to $\beta'$. Consequently, there is no need for an explicit representation of both $\alpha,\beta$ \textit{and} $\alpha',\beta'$ in the lifted graph. We formalize this reasoning in the next subsection.

\subsection {Symmetry-Preserving Properties}
The lifted GBP framework is based on two symmetry oriented properties. The first states that symmetry preserving permutations for the region graph are derived from the joint structure of the MRF and the set of chosen outer regions.
The second property formalizes the symmetry preserving nature of the flooding schedule, allowing for a compact representation of the message flow. Formally,
\begin{theorem}
Let $ \Omega$ denote a set of outer regions for MRF $\mathcal{M}$, and let $G_{\mathcal{R}}$ denote a region graph obtained by iterative intersections. Then, any permutation $\pi$ on $\mathcal{M}$'s variables which preserves the structure of both $\Omega$ and $\mathcal{M}$, induces a permutation $\pi_{_{\mathcal{R}}}$ which, when applied to the region graph's nodes, preserves its structure.
\end{theorem}
\begin{theorem}
\label{th:symMP}
Let $G$ denote a graph, and let $\pi_{_{\mathcal{R}}}$ be a structure preserving permutation on the graph's nodes. Let $\mathcal{A}$ denote a deterministic, graph structure based, message passing algorithm, applied to $G$ with a flooding schedule, and initialized such that for any pair of nodes $(u,v)$, the message from $u$ to $v$ (denoted by $m_{u \rightarrow v}$) is equal to $m_{\pi_{_{\mathcal{R}}}(u) \rightarrow \pi_{_{\mathcal{R}}}(v)}$ under the permutation $\pi_{_{\mathcal{R}}}$. Then, $m_{u \rightarrow  v}$ remains equal to $m_{\pi_{_{\mathcal{R}}}(u) \rightarrow \pi_{_{\mathcal{R}}}(v)}$ under $\pi_{_{\mathcal{R}}}$ throughout the entire execution of $\mathcal{A}$.
\end{theorem}

\begin{cor}
\label{cor:self_mapping}
Let region $\alpha$ denote the parent of regions $\beta_1$ and $\beta_2$ in the region graph, and let $\pi_{_{\mathcal{R}}}$ denote a structure preserving permutation, mapping $\beta_1$ to $\beta_2$ and $\alpha$ to itself. Then, all messages sent from $\alpha$ to $\beta_1$ are identical to those sent from $\alpha$ to $\beta_2$, under the permutation $\pi_{_{\mathcal{R}}}$.
\end{cor}
\subsection {Generating the Lifted Region Graph}
Before generating the lifted graph, we must choose which outer regions the represent. A natural pick would be all regions with scopes corresponding to factors in the ground MRF. The added value for this pick is that, if carefully constructed, each ground instance of a parfactor corresponds to a scope of some canonical cluster. More specifically, if a $X!=Y$ constraint is injected for each pair of lvars $X,Y$ in the model, and the model is shattered accordingly, the canonical outer regions can be picked without any computational effort.  
We continue with a formal definition of the lifted region graph.
\begin {defn}
A \textit{lifted region graph} $G_{\mathcal{R}}=(V,E,\sigma,\kappa)$ is a Directed Acyclic Multigraph, with nodes in $V$ denoting regions, directed edges in $E$ denoting parent to child relations, $\sigma$ denoting a function which associates each edge with a mapping from the parent's variables to those of the child, and $\kappa$ denoting a function which associates each edge with a positive integer, representing the cardinality of the parent-child relation. 
\end {defn}
A lifted region graph is obtained by identifying symmetrical intersections induced by the graph's regions, and representing each set of symmetrical regions via a single (canonical) region. The method for identifying these \textit{lifted} intersections is based on the following observation. Let the region graph consist of an edge $\alpha \rightarrow \beta$ between regions $\alpha$ and $\beta$. Then, for each structure preserving permutation $\pi_{_{\mathcal{R}}}$ mapping $\alpha$ to $\alpha'$, the region graph consists of an edge $\alpha' \rightarrow \beta'$, where $\beta'$ is the image of $\beta$ under $\pi_{_{\mathcal{R}}}$. Notably, intersections may exist between parent regions from different symmetry classes, or between parents of the same symmetry class. It is therefore convenient to approach this problem as a search over the canonical representations of all subsets in each of the regions.
\begin{algorithm} [t!]
\DontPrintSemicolon
\KwIn{Canonical representations of outer regions $\Omega$ \;
}
\KwOut{A lifted region graph $\GR$}
$V = \Omega $, $E=\{\}$, $\sigma = \{\}$, $\kappa = \{\}$ \;
$\GR \equiv (V,E,\sigma,\kappa)$ \;
\vspace{2pt}
Let $d_{max}$ denote the maximum size of any region in $\Omega$\;
\For {$d=d_{max}$ \bf{down to} $2$}  {
 	Let $D$ denote all regions of size $d$ in $V$\;
 	\For{$\alpha \in D$, $\beta \in$ subsets of $\alpha$ of size $d-1$}{
 		Let $\beta^*$ denote the canonical representation of $\beta$ \;
 		Add region $\beta^*$ to $V$\;
 		Let $\Sigma$ denote a set of mappings from $\beta$ to $\beta^*$, derived from isomorphisms of their CSGs\; 			
 		\For{$\varsigma \in \Sigma$}{
 			Let $k$ denote the cardinality of the $\alpha \rightarrow \beta^*$ parent-child relation under the mapping $\varsigma$\; 
 			Let $E$ consist of an edge $e$ from $\alpha$ to $\beta^*$, with properties $\sigma(e) = \varsigma$ and $\kappa(e) = k$}
 	}
}

\Return{$\GR$}
\caption{{ \sc GenerateLiftedRegionGraph}}
\label{algo:GenLiftedRegionGraph}
\end{algorithm}

\begin{figure}[t]
\centering
    \subfloat[Lifted Region Graph \label{fig:liftedRegionGraph}]{
      \includegraphics[clip, width=0.29\textwidth]{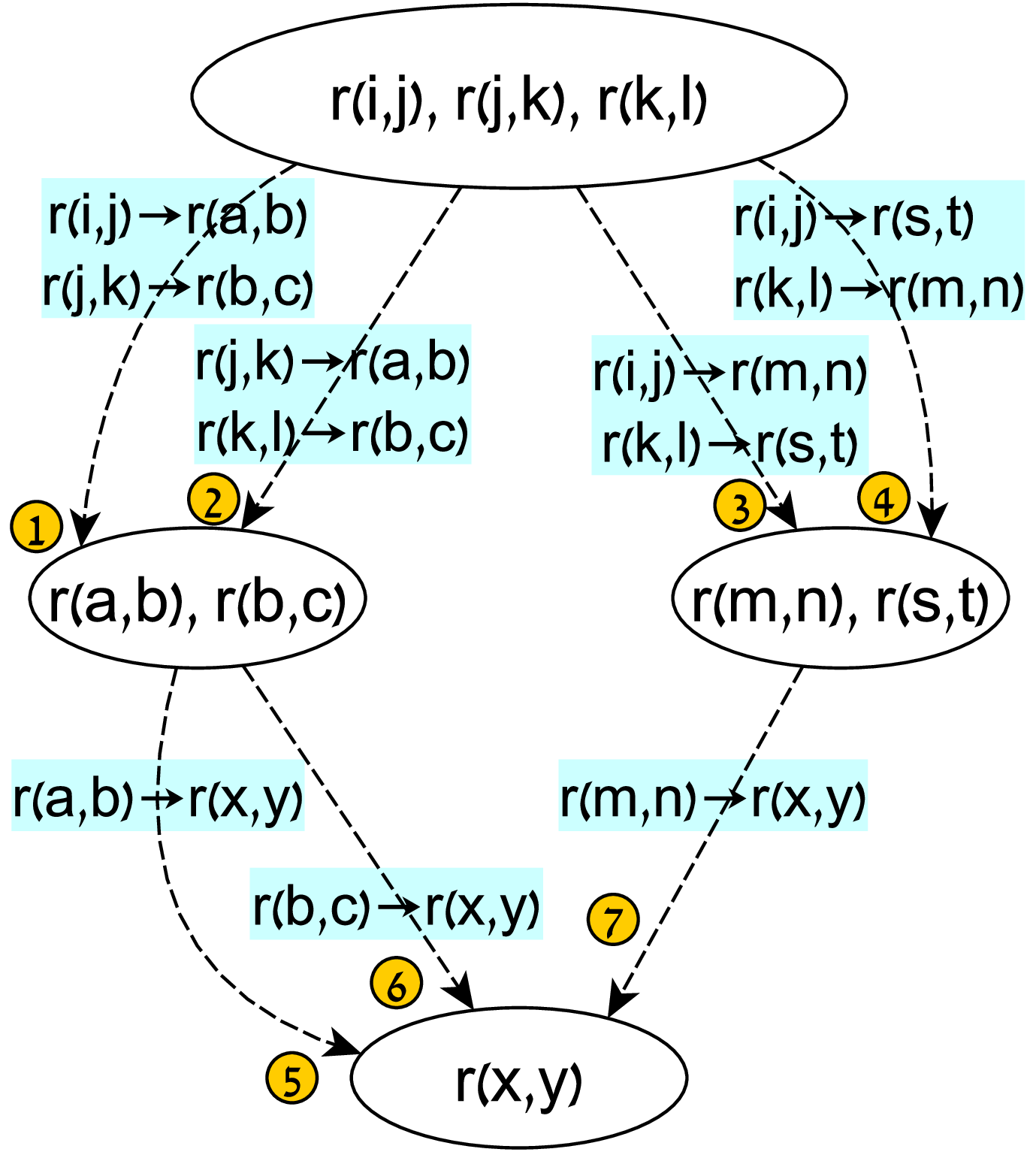}
     }
    \hspace {1pt}
    \subfloat[Cluster Signatures \label{fig:csg}]{
      \includegraphics[clip, width=0.275\textwidth]{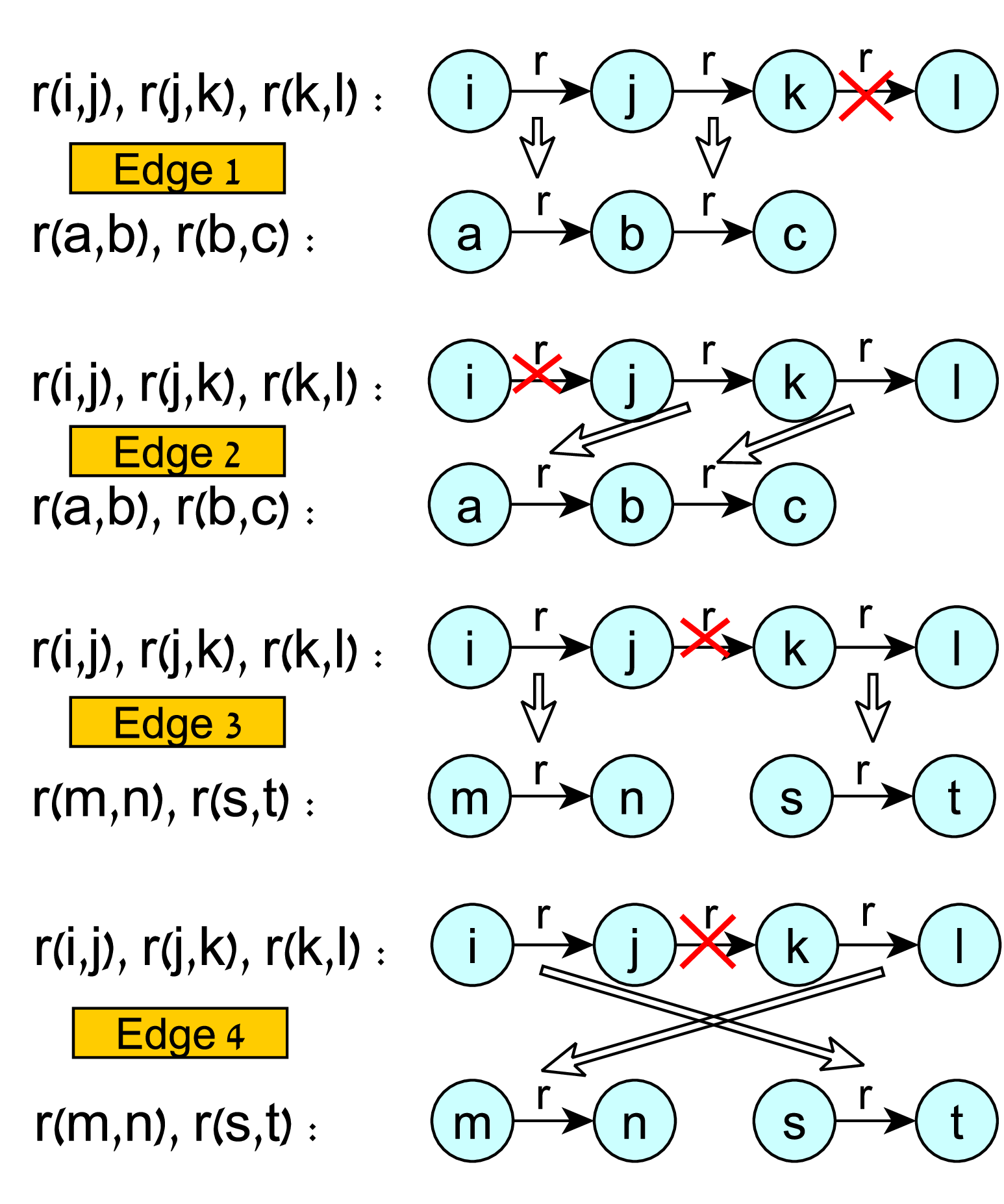}
    }
   \hspace {1pt}
    \subfloat[Local Graph of {\smaller $\{r_{(1,2)},r_{(2,3)},r_{(3,4)}\}$} \label{fig:localRegionGraph}]{
      \includegraphics[clip, width=0.38\textwidth]{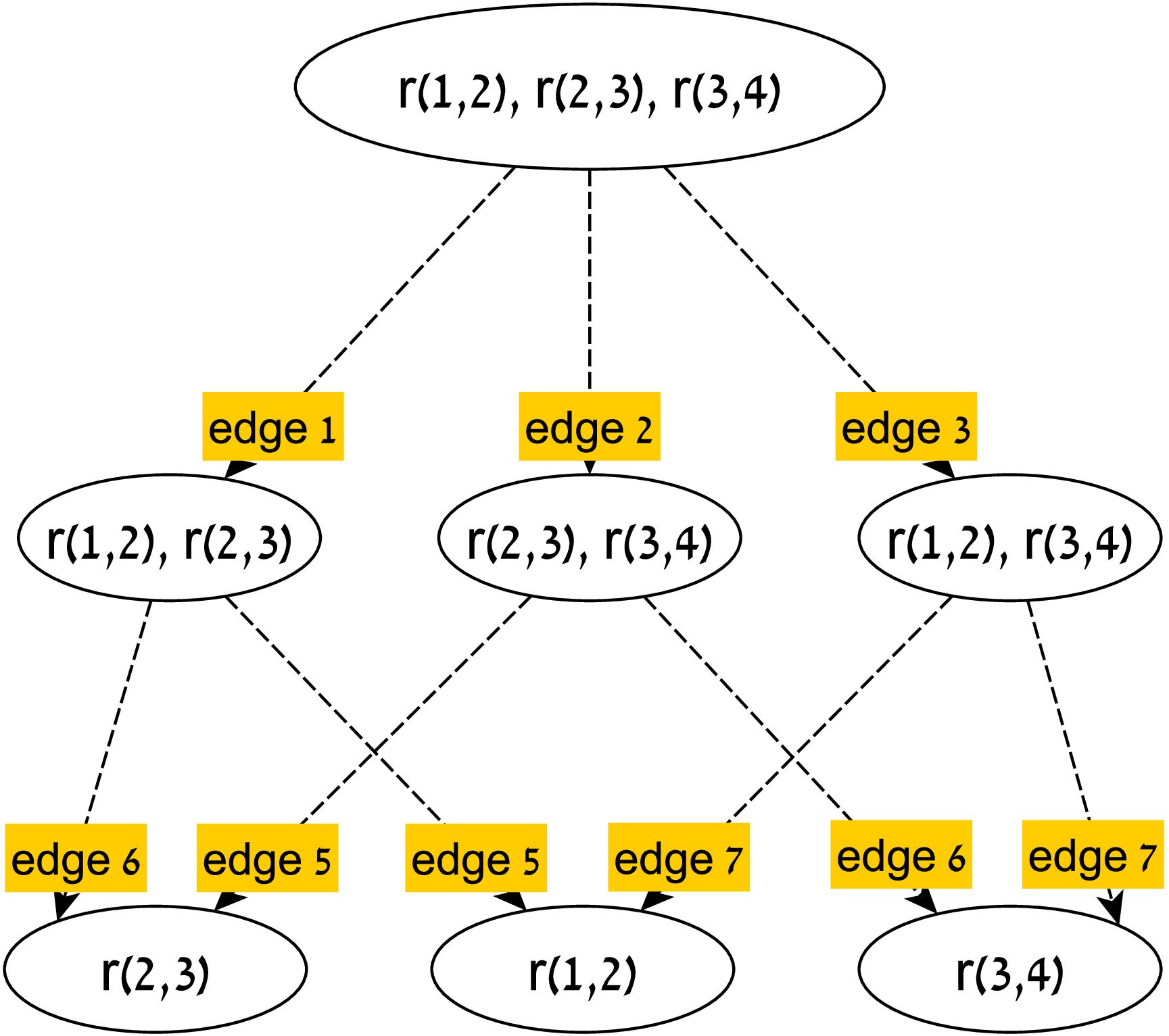}
     }
    \caption{Region Graphs and CSGs in {\small $\phi\big(r(X,Y),r(Y,Z),r(Z,W)\big)$ (all-distinct lvars constraint )}}
    \label{fig:RegionGraphs}
\end{figure}

Algorithm \ref{algo:GenLiftedRegionGraph} depicts the generation of the lifted region graph\footnotemark. The input of the algorithm is a set of indices of canonical clusters, which represent the outer regions. The output is the lifted region graph  $\GR$. The algorithm iterates over all subsets $\beta$ of size $d-1$ taken from regions of size $d$, starting with $d$ as the size of the maximal outer region, and down to 2. A canonical representation $\beta^*$ for each such subset is obtained and added as a child region to the canonical parent, under (potentially) several distinct directed edges: one for each mapping from $\beta$ to $\beta^*$, where the mappings are derived from all isomorphisms of their respective CSGs. See Figures \ref{fig:liftedRegionGraph} and \ref{fig:csg} for illustration.

However, some mappings from subsets of a canonical parent $\alpha$ to a canonical child $\beta^*$, must be filtered-out in order to maintain the lifted graph's correctness. If (not necessarily distinct) subsets of $\alpha$, denoted by $\beta_1$ and $\beta_2$, map onto the same $\beta^*$ and their mappings indicate the existence of a structure preserving permutation which maps $\beta_1$ to $\beta_2$ while mapping $\alpha$ to itself, then including both mappings as edges would result in over-counting the GBP messages. The existence of such a permutation can be detected via the following procedure. (i) Combine the mapping from $\beta_1$ to $\beta^*$ with the opposite mapping from $\beta_2$ to $\beta^*$, to produce a mapping from $\beta_1$ from $\beta_2$. (ii) The sought permutation exists iff the mapping from $\beta_1$ to $\beta_2$ is an automorphism of $\alpha$'s CSG. $p_i$ and $p_j$ from the canonical pair $\{p_i,p_j\}$ in Subsection \ref{subsec:overview}, are an example for such $\beta_1$ and $\beta_2$.
\footnotetext{We note that the procedure, under this chosen formulation, may produce a lifted region graph consisting of \textit{non-intersection} regions,  which serve as targets for only a single directed edge, and whose cardinality equals $1$. Such regions can alternatively be removed from the graph, by connecting their parent directly to their children.}
\subsection {Computing the Parent-Child Cardinality}
Let $\alpha \rightarrow \beta^*$ denote an edge from region $\alpha$ to $\beta^*$ in the lifted region graph. Let $\beta$ denote the subset of $\alpha$ mapped to $\beta^*$ under the edge's mapping. The cardinality of the edge, denote by $\kappa(\alpha \rightarrow \beta^*)$, is the number of all clusters in the relational MRF mapped to $\xI$ in a structure preserving permutation, while mapping all members of $\beta$ to themselves. Thus, for models with $N$ domain objects, where $n_{\beta}$ denotes the number of objects referenced in the cluster $\xJ$ and $n_{\alpha \setminus \beta}$ is the number of domain objects in $\xI$ not referenced by $\xJ$, the cardiniality is given by $
\kappa(\alpha \rightarrow \beta^*) = \prod_{i=1}^{n_{\alpha \setminus \beta}} N - n_{\beta} - i + 1$.

\subsection {Lifted Message Passing}
Messages in the lifted region graph are defined per a parent-to-child edge. Notably, pairs of parent-child, $\alpha$ and $\beta$, may be connected multiple times via distinct edges. The key to forming the messages is obtaining the belief states of both regions, $b_\alpha$ and $b_\beta$. Each belief is derived from both the structure of the lifted graph and the region's internal structure. The reason for this mixture is that different descendants of a region (e.g.~region $\alpha$), although might be members of the same symmetry class, may very well not be mapped one onto the other under any structure preserving permutation which maps $\alpha$ onto itself. Therefore, subsets of $\alpha$ which are symmetrical w.r.t.~the global structure, may still affect $\alpha$'s belief state in a non-symmetrical way. To overcome this obstacle, we define the notion of a \textit{local graph}. The local graph of canonical region $\alpha$ is the set of all its (non-canonized) subsets, connected such as to form a region graph whose most top region is $\alpha$ (see Figure \ref{fig:localRegionGraph}).

Before introducing the lifted messages, we define some related notations. $\sim x_{\alpha \rightarrow \beta}$ denotes all $\xI$ variables but those which take part in the mapping from $\alpha$ to $\beta$. $inEdges(\alpha)$ denotes all $\alpha$'s incoming edges in the lifted region graph, where each such edge will be denoted by $\rho \rightarrow \alpha$. $localDesc(\alpha)$ denotes all non-canonized subsets of $\alpha$, $\gamma^*$ denotes the canonical representation of a region $\gamma$, and $localPar(\alpha,\gamma,\rho^* \rightarrow \gamma^*)$ denotes all members of $\alpha$'s local graph (including $\alpha$) that are parents of $\gamma$, and are connected via an edge in the local graph whose corresponding edge in the lifted region graph is $\rho^* \rightarrow \gamma^*$. Lastly, the notation $\sigma_{(\alpha \rightarrow \beta)}: \chi$ denotes the result of applying the mapping associated with the lifted edge $\alpha \rightarrow \beta$, to the variables of function $\chi$. The update rules of the lifted message passing are given by the following equations.
\begin{align}
&b_\alpha (\xI) \propto  \hspace{19pt}
\prod_{f \in f_\alpha}{\phi_f(x_f)} 
\prod_{(\rho \rightarrow \alpha) \in inEdges(\alpha)} { m_{\rho \rightarrow \alpha} (\xI) } ^{\; \kappa(\rho \rightarrow \alpha)} \\
&\prod_{\gamma \in localDesc(\alpha)}  \; 
\sigma_{(\gamma^* \mapsto \gamma)}:
\prod_{(\rho^* \rightarrow \gamma^*) \in inEdges(\gamma^*)}{m_{\rho^* \rightarrow \gamma^*}(x_{\gamma^*}) } 
^{\; \kappa(\rho^* \rightarrow \gamma^*) - |localPar(\alpha,\gamma,\rho^* \rightarrow \gamma^*)|
} \nonumber
\end{align}
\begin{equation}
m_{\alpha \rightarrow \beta}(x_\beta) \longleftarrow 
\frac{\sigma_{(\alpha \rightarrow \beta)} : 
\sum_{\sim x_{\alpha \rightarrow \beta}} {b_\alpha(x_\alpha)} } {b_\beta(x_\beta)} \; m_{\alpha \rightarrow \beta}(x_\beta)
\end{equation}
Finding $localPar(\alpha,\gamma,\rho^* \rightarrow \gamma^*)$ can be non-trivial. One way to obtain this set of local regions is to "reconstruct" the procedure that associated $\gamma^*$ with all its lifted parents, as follows. Let $\sigma_{(\gamma^* \mapsto \gamma)}$ denote a mapping from $\gamma^*$ to $\gamma$, derived from an isomorphism of their CSGs. We will define two sets of graphs $\mathcal{G}$ and $\mathcal{G}^*$ such that the combination of isomorphisms from members of $\mathcal{G}$ to $\mathcal{G}^*$ guarantees adequate matchings from all edges in $\alpha$'s local graph to those in the lifted graph. A distinct integer is associated with each member of $\gamma$, $1$ through $|\gamma|$. Members of $\mathcal{G}$ are defined as the CSGs of all parents of $\gamma$ in the local graph, modified such that edge colors of members of $\gamma$ (e.g.~'$r$','$p$'), are augmented by the associated integer (e.g.~'$r=1$','$p=3$'). 

$\mathcal{G}^*$ consists of one graph per each edge terminating in $\gamma^*$ in the lifted region graph. Each such graph is a CSG with modified colors. More specifically, let $\rho^* \rightarrow \gamma^*$ denote an edge in the lifted region graph whose associated mapping is $\sigma_{(\rho^* \rightarrow \gamma^*)}$. $\mathcal{G}^*$ should then consist of the CSG of $\rho^*$, with colors that are modified according to a matching from the members of $\rho^*$ to the members of $\gamma$. Matching from $\rho^*$ to $\gamma$ is done by combining the mapping $\sigma_{(\rho* \rightarrow \gamma^*)}$ with $\sigma_{(\gamma^* \mapsto \gamma)}$. If the mapped $\gamma$ member consists of the color '$r=1$', so will the corresponding $\rho^*$ member. Lastly, a graph in $\mathcal{G}$ that is isomorphic with a graph in $\mathcal{G}^*$, will be associated with the latter's edge. Figure \ref{fig:localRegionGraph} depicts the association of edges in a local graph obtained from the lifted region graph in Figure \ref{fig:liftedRegionGraph}.

 \begin{figure}[t]
\centering
    \subfloat[Running Times \label{fig:times}]{
      \includegraphics[clip, width=0.49\textwidth]{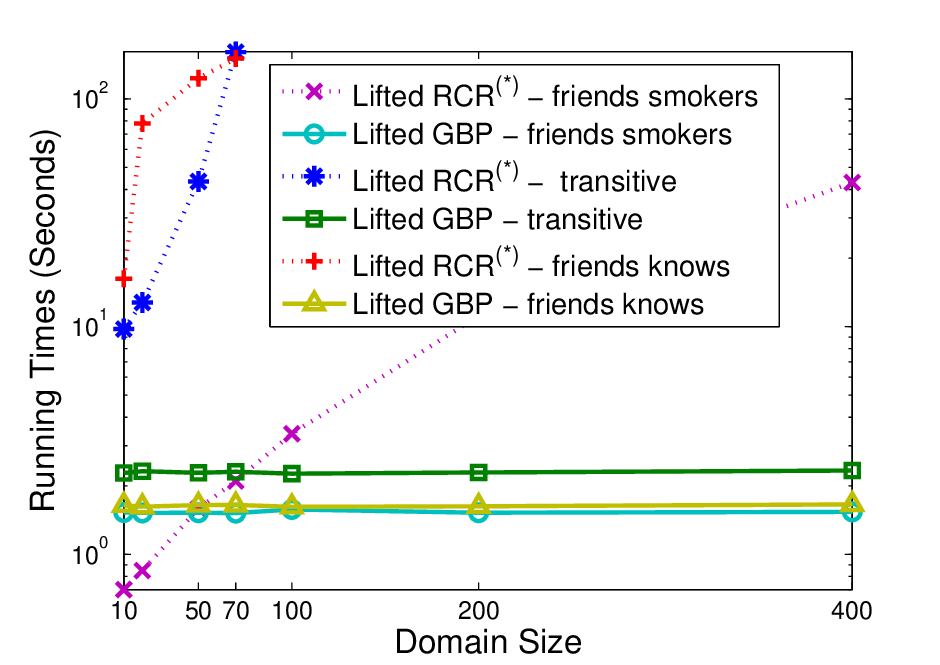}
     }
    \subfloat[Query Probability of $friends(X,Y)$ \label{fig:probs}]{
      \includegraphics[clip, width=0.49\textwidth]{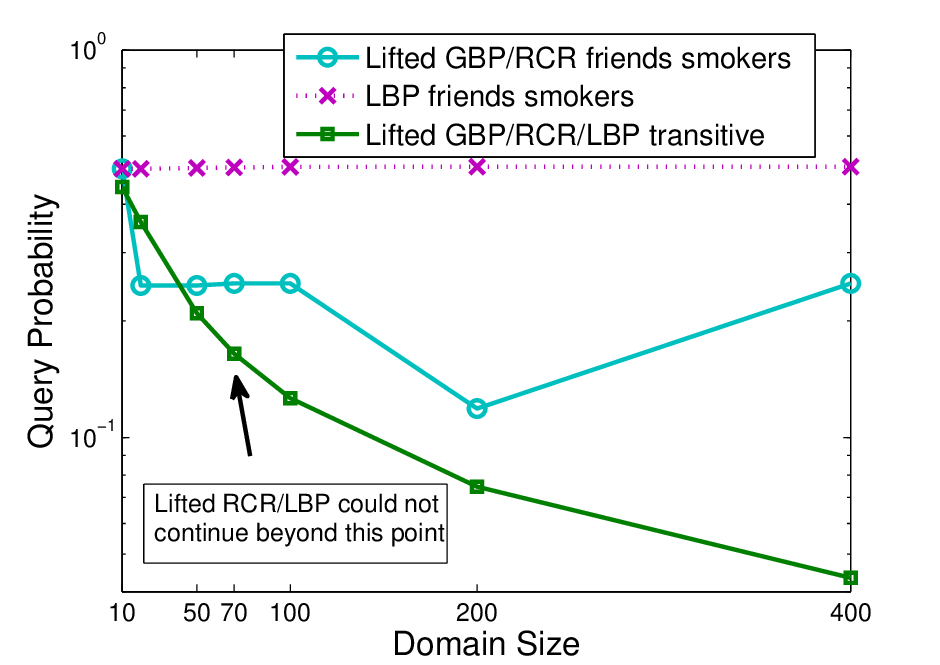}
    }
    \caption{Empirical Results}
    \label{fig:results}
\end{figure}

\section {Empirical Evaluation}
We conducted experiments on $4$ structurally distinctive relational models. The first model is a \textit{friends smokers} model  (i.e. $friends(X,Y) \wedge smokes(X) \implies smokes(Y)$), the second is a \textit{transitive model} \cite{apsel14AAAI}, the third is a transitive \textit{friends knows model} \cite{DBLP:conf/uai/ApselB12}, and the forth is the \textit{chain model} whose region graphs are depicted in Figure \ref{fig:RegionGraphs}. Graph isomorphism tests were conducted using the NetworkX package \cite{hagberg2008exploring}. The time required to generate the lifted graphs turned out to be in the range of tens of milliseconds. In all experiments, we ran 500 iterations of lifted GBP message passing with a damping factor of 0.5 and normalization of both beliefs and messages. Message computations were carried in log space in order to maintain numerical stability. Time performance and query results were compared against the WFOMC \cite{DBLP:conf/ijcai/BroeckTMDR11} engine, which is capable of computing GBP queries (\textit{Lifted RCR}), as well as lifted BP (\textit{LBP}). Lifted RCR times were normalized by reducing the time of respective LBP computations, thereby eliminating engine overhead of parsing and preprocessing from the comparison. Since our implementation is lightweight and domain size independent, it significantly dominates in time performance. Results of probabilistic queries were practically identical for RCR and our implementation. Notably, non-generalized lifted BP computations for the transitive model, return very similar results to GBP, and were omitted. The chain model, which does not appear in any of the figures, did not converge neither in RCR nor in our implementation.
\section {Conclusions}
We introduced a lifted GBP message passing algorithm which is domain-size independent in shattered models. Although the scope of this work is confined to sum-product queries, the compact formulation of the lifted region graph may serve other algorithms which rely on a similar structure. One aspect worth exploring is extending the scope of outer regions, not only to regions defined using first-order theory, but also to all regions of a given size, similarly to lift and project hierarchies \cite{DBLP:journals/siamdm/SheraliA90}. Such attempts may lead to tightened results and improved convergence in highly complex models.

\newpage
\bibliographystyle{abbrv}
\bibliography{LiftedGBP}

\end{document}